\documentclass[conference]{IEEEtran}
\IEEEoverridecommandlockouts
\usepackage{amsmath,amssymb,amsfonts}
\usepackage{algorithmic}
\usepackage{graphicx}
\usepackage{textcomp}
\usepackage{xcolor}
\def\BibTeX{{\rm B\kern-.05em{\sc i\kern-.025em b}\kern-.08em
    T\kern-.1667em\lower.7ex\hbox{E}\kern-.125emX}}

\usepackage{makecell}

\usepackage[
backend=biber,
style=ieee,
sorting=ynt
]{biblatex}
\usepackage{hyperref}

\begin{document}

\title{Deep Architectures for \\Content Moderation and Movie Content Rating}
\author{\IEEEauthorblockN{Fatih Cagatay Akyon}
\IEEEauthorblockA{ \\
\textit{Graduate School of Informatics, METU}\\
Ankara, Turkey \\
}
\and
\IEEEauthorblockN{Alptekin Temizel}
\IEEEauthorblockA{ \\
\textit{Graduate School of Informatics, METU}\\
Ankara, Turkey \\
}
}

\maketitle

\begin{abstract}
Rating a video based on its content is an important step for classifying video age categories. Movie content rating and TV show rating are the two most common rating systems established by professional committees. However, manually reviewing and evaluating scene/film content by a committee is a tedious work and it becomes increasingly difficult with the ever-growing amount of online video content. As such, a desirable solution is to use computer vision based video content analysis techniques to automate the evaluation process. In this paper, related works are summarized for content moderation, video classification, multi-modal learning, and movie content rating. The project page is available at \href{https://github.com/fcakyon/content-moderation-deep-learning}{https://github.com/fcakyon/content-moderation-deep-learning}.
\end{abstract}


\section{Introduction}
Many films including movies, TV series, animations, and other audio-visual content have inappropriate words or visual content unsuitable for uncongenial audiences and adversely affects the younger generation. Primarily, it affects under-aged audiences through their exposure to visual content on TV or other video-sharing and social networking websites like YouTube, Facebook. Hence, it is desirable to automatically detect sensitive content for filtering, making Automated Film Censorship and Rating (AFCR) one of the prominent applications of Machine Learning (ML) and attracting research. 

Identification of sensitive content is a challenging task due to being subjective and open ended \cite{moreira2019multimodal} \cite{khaksar2021survey}. While broadcasters take proactive steps to identify sensitive content manually, this is a slow, monotonous, and inefficient process \cite{lyn2020no}. As a result, AFCR is required to generate automatic and aggregated ratings for films or any audiovisual content to determine the minimum age group suitable for viewing the content. Various aspects of a film that may be considered inappropriate include violence, nudity, substance abuse, profanity, abnormal psychological content, or other topics inappropriate for children or teenagers in general \cite{gajula2020machine}. Based on this content, films or any audiovisual content prepared for public display in many developed countries are subject to appropriate classification standards.

Any audio-visual content is broadly categorized into three parts: advisory, restricted, and adult or nudity. The advisory categories are further divided into General (G), Parental Guidance (PG), and Mature (M). The restricted types include Mature Accompanied (MA15+) and Restricted (R18+). All these classifications have specified minimum age for viewing any audio-visual content under that specific class. ML plays a vital role in film censorship to detect and classify sensitive and inappropriate content into a given classification standard \cite{motion2010classification} \cite{afsha2022machine} \cite{khaksar2021survey}. Most researchers utilized ML techniques such as Convolutional Neural Network (CNN) \cite{haque2022detection} \cite{chai2021automatic} \cite{pandey2021device} \cite{larocque2021gore} \cite{dhanwal2020automated}, Support Vector Machine (SVM) \cite{moreira2019multimodal} \cite{de2019multimodal} \cite{de2019baseline}, Long ShortTerm Memory (LSTM) \cite{chai2021automatic} \cite{gunawan2020automatic}, Gated Recurrent Units (GRU) \cite{haque2022detection}, Vision transformers \cite{chen2022movies2scenes} in their work for detecting unsuitable audio-visual content.

\section{Related Work}

This section of this paper aims to provide a brief overview of existing research on the five main topics addressed in this study. Specifically, we will discuss previous studies on content moderation and movie datasets, action recognition, multi-modal learning, movie genre classification, and sensitive content detection, highlighting their key findings and contributions to the field. This literature review will provide a context for multi-modal content rating research.

\subsection{Content Moderation and Movie Datasets}

\begin{table*}
\small
\begin{center}
\caption{List of content moderation and movie datasets.}
\label{table:datasets}
\vspace{0.3cm}
\begin{tabular}{ c c c c c } 
\hline
Name & Year & Modality & Task & Labels \\
\hline
LSPD \cite{LSPD_dataset} & 2022 & image, video & \makecell{image/video classification, \\ instance segmentation}& \makecell{porn, normal, sexy, hentai, \\ drawings, female/male genital, \\ female breast, anus} \\ 
\hline
MM-Trailer \cite{mmtrailer_dataset} & 2021 & video & video classification & age rating \\
\hline
Movienet \cite{movienet_dataset} & 2021 & image, video, text &  \makecell{object detection, \\ video classification} & \makecell{scene level actions and places, \\ character bboxes} \\
\hline
\makecell{Movie script \\ severity dataset \cite{movie_scripts_severity_dataset}} & 2021 & text &  \makecell{text classification} & \makecell{frightening, mild, \\ moderate, severe} \\
\hline
LVU \cite{lvu_dataset} & 2021 & video & video classification & \makecell{relationship, place, like ration, \\ view count, genre, writer, year \\ per movie scene} \\
\hline
\makecell{Violence detection \\ dataset \cite{violence_det_dataset}} & 2020 & video & video classification & violent, not-violent \\
\hline
\makecell{Movie script \\ dataset \cite{movie_script_dataset}} & 2019 & text & text classification & violent or not \\
\hline
\makecell{Adult content \\ dataset \cite{adult_content_dataset}} & 2017 & image & image classification & nude or not \\
\hline
\makecell{Substance use \\ dataset \cite{substance_use_dataset}} & 2017 & image & image classification & drug related or not \\
\hline
\makecell{NDPI 2k \cite{ndpi_2k_dataset_trof}} & 2016 & video & video classification & porn or not \\
\hline
\makecell{Violent Scenes \\ Dataset \cite{violent_scenes_15_dataset}} & 2014 & video & video classification & blood, fire, gun, gore, fight \\
\hline
VSD2014 \cite{violent_scenes_14_dataset} & 2014 & video & video classification & blood, fire, gun, gore, fight \\
\hline
AIIA-PID4 \cite{aiia_pid4_dataset} & 2013 & image & image classification & bikini, porn, skin, non-skin \\
\hline
NDPI 800 \cite{ndpi_800_dataset} & 2013 & video & video classification & porn or not \\
\hline
HMDB-51 \cite{hmdb51_dataset} & 2011 & video & video classification & smoke, drink \\
\hline
\end{tabular}

\end{center}
\end{table*}

There are many published datasets in the context of content moderation and movies. Substance use dataset \cite{substance_use_dataset} focuses on drug detection; however, it is not publicly available. Publicly available HMDB-51 action recognition dataset \cite{hmdb51_dataset} is loosely related to the field of video content moderation as it contains smoking and drinking classes which can be considered as substance use. VSD2014 \cite{violent_scenes_14_dataset} and Violent Scenes Dataset \cite{violent_scenes_15_dataset} provide violence-related labels for Hollywood movie scenes and annotations are publicly available. NDPI 800 \cite{ndpi_800_dataset}, AIIA-PID4 \cite{aiia_pid4_dataset}, NDPI 2k \cite{ndpi_2k_dataset_trof}, Adult content dataset \cite{adult_content_dataset} and LSPD \cite{LSPD_dataset} are image and video based datasets that provide nudity and pornography detection related annotations which are available to researchers upon request. Movie script dataset \cite{movie_script_dataset} and Movie script severity dataset \cite{movie_scripts_severity_dataset} are text (subtitle and caption) based datasets that focus on severity and violence detection. Lastly, LVU \cite{lvu_dataset}, MovieNet \cite{movienet_dataset} and MM-Trailer \cite{mmtrailer_dataset} are movie-related datasets that contain character level relationship, background place, character bounding boxes, and age rating per trailer however they do not contain any rating annotations. More information on the datasets are provided in Table \ref{table:datasets}.

\subsection{Action Recognition}

Human action recognition is an active research area and it plays a significant role in video understanding. Prior action recognition works focus on 3D CNN based architectures \cite{i3d} \cite{r2p1d} \cite{slowfast} \cite{x3d}. A two stream approach where optical flow and frame sequence inputs are utilized is proposed in \cite{i3d}. R(2+1)D convolutions, which decomposes the 3D convolution into a 2D spatial convolution and a 1D temporal convolution is used in \cite{r2p1d}. In \cite{slowfast}, a two-stream approach is proposed. It simultaneously embeds slow and fast sampled frame sequences from the target video clip and fuses the features extracted from both paths. In \cite{x3d}, a family of efficient video networks which progressively expand a tiny 2D image classification architecture along multiple network axes, in space, time, width, and depth are presented. At each step, one axis is expanded after training and validation.

After the rise of vision transformers, many transformer-based action recognition, and video classification techniques have been proposed \cite{timesformer} \cite{video-swin} \cite{bevt} \cite{videomae}, which compare favorably against 3D CNN based models. Since transformers have no inductive bias about the nature of the data, they can scale better compared to CNN based encoders provided that large training data is available. TimeSformer \cite{timesformer} extends the standard vision transformer (ViT) \cite{vit} with the addition of temporal transformer encoder layers on top of a spatial encoder to encode spatio-temporal information. Furthermore, Video Swin Transformer \cite{video-swin} extends Swin Transformer \cite{swin} by integrating an inductive bias of locality in videos utilizing the pretrained weights of Swin Transformer. BEVT \cite{bevt} studies BERT pretraining for video transformers and proposes masked pretraining \cite{videomae} of the transformer encoder and holds the best representation learning capabilities as of today.

These methods can be utilized in any video classification and video representation learning problems such as movie scene content rating.

\subsection{Multi-Modal Learning}

\begin{table*}
\small
\begin{center}
\caption{Summary of video based content moderation techniques.}
\label{table:vid_cont_mod}
\vspace{0.3cm}
\begin{tabular}{ c c c c c c c } 
\hline
Paper & Year & Model & Features & Datasets & Tasks & Context \\
\hline
\cite{son2022reliable} & 2021 & \makecell{separate model\\ for each task: \\ concat + LSTM, \\ object detector, \\ one-class CNN \\ embeddings}& \makecell{video frame \\ pixel values, \\ image \\ embeddings, \\ text} & \makecell{Nudenet, \\ private \\ dataset} & \makecell{profanity, violence, \\ nudity, \\ drug classification} & \makecell{movie \\ content rating} \\
\hline
\cite{dhanwal2020automated} & 2020 & CNN + MLP & \makecell{InceptionV3 \\ image \\ embeddings} & \makecell{private dataset} & \makecell{cigarette \\ classification \\ from video} & \makecell{general \\ video \\ content \\ moderation} \\
\hline
\cite{de2019multimodal} & 2019 & \makecell{CNN + SVM} & \makecell{InceptionV3 \\ image \\ embeddings, \\ AudioVGG \\ audio \\ embeddings} & \makecell{private dataset} & \makecell{inappropriate \\ (nudity+gore) \\ classification \\ from video} & \makecell{general \\ video \\ content \\ moderation} \\
\hline
\cite{de2019baseline} & 2019 & \makecell{concat. features + \\ SVM, MLP}& \makecell{InceptionV3 \\ image \\ embeddings, \\ AudioVGG \\ audio \\ embeddings} & \makecell{YouTube8M, \\ NDPI 800, \\ Cholec80} & \makecell{nudity \\ classification \\ from video} & \makecell{e-learning \\ content \\ moderation} \\
\hline
\cite{tahir2019bringing} & 2019 & \makecell{CNN + LSTM \\ (late fusion)}& \makecell{CNN based \\ encoder for \\ image, video \\ and audio \\ spectrograms} & private dataset & \makecell{video \\ (classification: \\ orignal, \\ fake explicit, \\ fake violent} & \makecell{social media \\ content \\ moderation} \\
\hline
\end{tabular}

\end{center}
\end{table*}

With the increase of available multi-modal data (mainly image, video, text modalities) and transformer-based encoders that can unify different modalities into similar embedding space, research around multi-modal neural networks has accelerated. Multi-modal architectures can be grouped into two main categories: asynchronous and synchronous models.

When different input modalities that are used in a training step of a model do not align in time, this model is called an asynchronous multi-modal architecture. In the training of these models, independent image/video/text based datasets can be utilized. PolyViT \cite{likhosherstov2021polyvit} proposes the usage of ViT-based encoder for image, audio, and video modalities and proposes a multi-task based pretraining with separate classification head for each dataset (with image/video/audio classification tasks). Similarly, Omnivore \cite{girdhar2022omnivore} proposes a multi-task based pretraining schema training a model on image/video classification and depth estimation. OmniMAE \cite{girdhar2022omnimae} also proposes a ViT-based encoder but instead of supervised pretraining a self-supervised masked pretraining is proposed for image and video-based datasets. Then they fine-tune the pretrained model for downstream tasks as image and video classification.

When a model requires different input modalities of the same instance to be aligned in time, this model is called synchronous multi-modal architecture. In the training of these models, modalities should be aligned and cannot be regarded as independent. There are works that use image + text \cite{lu202012} \cite{chen2020uniter} \cite{openai_clip}, video + audio \cite{morgado2021robust} \cite{morgado2021audio}, video + optical flow \cite{simonyan2014two}, video + text \cite{ni2022expanding}, image + depth map \cite{bachmann2022multimae}, image + audio + text \cite{akbari2021vatt} \cite{baevski2022data2vec}, image + audio + optical flow \cite{xiong2022m} modalities to learn more representative multi-modal embeddings for classification, depth estimation, image retrieval and pretraining tasks. Some of these works propose the usage of separate encoders per modality \cite{simonyan2014two} \cite{openai_clip} \cite{morgado2021robust} \cite{morgado2021audio} \cite{ni2022expanding} \cite{akbari2021vatt} while others propose a single unified encoder \cite{bachmann2022multimae} \cite{xiong2022m} \cite{chen2020uniter}. Earlier works used CNN based embedding extractors for image and video modalities \cite{simonyan2014two} while all previously mentioned other methods utilized transformer-based encoders.

\subsection{Movie Genre Classification}

\begin{table*}
\small
\begin{center}
\caption{Summary of video based movie content rating techniques.}
\label{table:vid_mov_rating}
\vspace{0.3cm}
\begin{tabular}{ c c c c c c c } 
\hline
Paper & Year & Model & Features & Datasets & Tasks & Content Rating Context \\
\hline
\cite{chen2022movies2scenes} & 2022 & \makecell{ViT-like \\ video encoder \\ + MLP} & \makecell{ViT-like \\video encoder \\embedings} & Private & \makecell{ movie scene \\ representation learning, \\ video classification \\ (sex, violence, \\ drug-use)} & \makecell{movie scene} \\
\hline
\cite{haque2022detection} & 2022 & \makecell{CNN + GRU \\ + MLP} & \makecell{CNN \\ embeddings \\ from video \\ frames} & \makecell{Violence \\ detection \\ dataset} & \makecell{violent/non-violent \\ classification from \\ videos} & \makecell{movie scene} \\
\hline
\cite{shafaei2021case} & 2021 & \makecell{multi-modal \\ + multi output \\ concat. + MLP}& \makecell{CNN+LSTM \\ video features, \\ BERT text \\ embeddings, \\ MFCC audio \\ features} & MM-Trailer & \makecell{rating classification \\ (red, yellow, green)} & \makecell{movie trailer} \\
\hline
\cite{gunawan2020automatic} & 2020 & \makecell{3 CNN \\ model for 3 \\ modality, \\ multi-label \\ dataset}& \makecell{CNN video \\ and audio \\ embeddings, \\ LSTM text \\ (subtitle) \\ embeddings} & \makecell{private dataset} & \makecell{gore, nudity, drug, \\ profanity classification \\ from video and subtitle} & \makecell{movie scene} \\
\hline
\cite{moreira2019multimodal} & 2019 & \makecell{meta- \\ learning \\ with Naive \\ Bayes, SVM} & \makecell{MFCC and \\ prosodic \\ features \\ from audio, \\ HoG and \\ TRoF features \\ from images} & \makecell{NDPI 2k \\ dataset, \\ VSD2014} & \makecell{violent and pornographic \\ scene localization from \\ video} & \makecell{movie scene} \\
\hline
\end{tabular}

\end{center}
\end{table*}

The movie genre plays an important role in video analysis, reflecting narrative elements, aesthetic approaches, and emotional responses. Availability of a robust method for video genre classification enables a wide range of applications, such as organizing related user videos from social networking sites, fixing mislabeled videos, extracting keyframes from long videos, and searching for a specific movie type for recommender systems. Motivated by these applications, researchers have proposed different techniques \cite{liu2020towards} \cite{yadav2020unified} \cite{shambharkar2020genre} \cite{mangolin2020multimodal} \cite{rajput2022multi} \cite{turkoz2022detection} \cite{liu2022msl} \cite{zhang2022effectively} for genre classification. \cite{liu2020towards} performs story-based classification for movie scenes using manually extracted categorical features in a logistic regression based model. Some methods utilize only the video frames while performing genre classification. \cite{yadav2020unified} uses Inception v4 \cite{inceptionv4} image embeddings in a CNN + LSTM based model, \cite{shambharkar2020genre} utilizes a 3D CNN based architecture to extract spatio-temporal features from video frames, \cite{turkoz2022detection} uses EfficientNet \cite{efficientnet} image embeddings in a CNN + GRU based model for genre classification from movie trailers. \cite{rajput2022multi} proposes a KNN based multi-label movie genre classification technique that uses only text frequency vectors extracted from trailer subtitles. There are some multi-modal approaches for genre classification. \cite{mangolin2020multimodal} proposes a technique that uses MFCC \cite{mfcc} and Local Binary Pattern (LBP) features of the audio, LBP and 3D CNN features extracted from video frames, Inception v3 \cite{inceptionv3} embeddings of poster images and Term Frequency-Inverse Document Frequency (TF-IDF) features of movie synopsis for multi-label movie genre classification. This work proposes to train separate model for each input feature and target label which makes the deployment and scaling harder. \cite{liu2022msl} proposes a novel transformer-based architecture utilizing Resnet18 \cite{resnet} image embeddings and Resnet-VLAD \cite{resnetvlad} audio embeddings for single-label news scene segmentation/classification. \cite{zhang2022effectively} proposes fusion of pretrained multi-modal CLIP \cite{openai_clip} text and image embeddings and PANN audio embedding for movie genre classification.

\subsection{Sensitive Content Detection}

Many movies, TV series, animations and other audio-visual content, contain inappropriate words or visual content that is unsuitable for particular audiences and younger generation. On the other hand, in online platforms or in personal devices there may be a need for filtering out irrelevant, obscene, illegal, harmful, or insulting content to emphasize useful or informative samples. With these motivations, there have been many ML based techniques proposed for the identification of sensitive content. These can be grouped into two categories as movie content rating and content moderation. Although target categories and motivations are similar between these groups, content moderation deals with more open domain data where movie content rating focuses on movies, TV series, and trailers.

There are some works that focus on image based content moderation \cite{larocque2021gore} \cite{pandey2021device}. \cite{larocque2021gore} proposes ensemble of CNN networks using image embeddings extracted with Mobilenet v2 \cite{mobilenetv2}, Densenet \cite{densenet} and VGG16 \cite{vgg16} backbones for gore classification from images using a private dataset. \cite{pandey2021device} proposes a network that performs SSD \cite{ssd} based object detection using Mobilenet v3 \cite{mobilenetv3} backbone for on-device person detection and nudity classification. The source code and dataset are not available.
Moreover, some works focus on video-based content moderation \cite{tahir2019bringing} \cite{de2019baseline} \cite{de2019multimodal} \cite{dhanwal2020automated}. \cite{tahir2019bringing} uses CNN-based encoder for video frames and audio spectrograms and after late fusion LSTM stage is applied to perform video classification for social media content moderation. \cite{de2019baseline} proposes to extract image embeddings from Inception v3 \cite{inceptionv3} encoder and audio embeddings from AudioVGG \cite{audiovgg} encoder and training with the concatenated features for nudity classification from e-learning videos. They use YouTube8M \cite{youtube8m_dataset}, NDPI 800 \cite{ndpi_800_dataset} and Cholec80 \cite{cholec80_dataset} datasets in their work. Similarly, \cite{de2019multimodal} uses Inception v3 \cite{inceptionv3} image embeddings and AudioVGG \cite{audiovgg} audio embeddings on an SVM for gore and nudity classification from video. \cite{dhanwal2020automated} proposes a method for cigarette classification from video using Inception v3 \cite{inceptionv3} image embeddings on a CNN based model. Summary of the video-based content moderation papers can be seen in Table \ref{table:vid_cont_mod}.

Earlier movie content rating works focus on image based rating \cite{tofa2017inappropriate} \cite{vishwakarma2019hybrid}. \cite{tofa2017inappropriate} proposes a technique consisting of SVM classifier, CNN image classifier and rule-based decision maker utilizing HoG and CNN features extracted from images for rating movie frames in terms of nudity and violence. \cite{vishwakarma2019hybrid} uses DCT \cite{DCT} features extracted from images to train an SVM classifier for movie content rating classification. Datasets used in these papers are not available.
Furthermore, there are more comprehensive works that focus on multi-modal movie content rating by utilizing video and/or text data \cite{gruosso2019deep} \cite{moreira2019multimodal} \cite{gunawan2020automatic} \cite{mmtrailer_dataset} \cite{movie_scripts_severity_dataset} \cite{chai2021automatic} \cite{haque2022detection} \cite{chen2022movies2scenes}. \cite{gruosso2019deep} proposes a rule based violent scene detection method utilizing Inception v3 \cite{inceptionv3} image embeddings. They use Violent Scenes Dataset \cite{violent_scenes_15_dataset} and a private dataset in their work. \cite{moreira2019multimodal} proposes a method for detecting violent and pornographic scenes using MFCC \cite{mfcc} and prosodic features from audio and HOG \cite{hog} and TRoF \cite{ndpi_2k_dataset_trof} features from images on a meta-learning based training schema. \cite{gunawan2020automatic} proposes a multi-label movie scene content rating technique that contains 3 different CNN-based model for 3 different modalities (audio, video, text). Related dataset is not available. \cite{mmtrailer_dataset} proposes a multi-modal architecture that utilizes CNN+LSTM spatio-temporal video features, BERT \cite{bert} and Deepmoji \cite{deepmoji} text features and MFCC audio features for rating movie trailers. Their dataset is available upon request. \cite{movie_scripts_severity_dataset} proposes a multi-task pairwise ranking \& classification network utilizing GloVe \cite{glove}, BERT and TextCNN \cite{textcnn} text embeddings of the movie scripts to predict the movie script severity. Their dataset is publicly available. \cite{chai2021automatic} aims to solve various rating tasks such as profanity, violence, nudity, drug classification but proposes separate network per task utilizing raw video frame pixel values, image and text embeddings. Most of their dataset is private and not available. \cite{haque2022detection} proposes CNN and GRU based violent scene detection network using the Violent Scenes Dataset \cite{violent_scenes_15_dataset}. None of the aforementioned methods utilize attention-based feature extraction and fusion methods and more recent vision and audio transformers such as ViT \cite{vit}, DEiT \cite{deit}, Wav2Vec \cite{wav2vec2}, Data2Vec \cite{baevski2022data2vec}. Unlike them, \cite{chen2022movies2scenes} proposes a movie scene representation learning method for ViT-like video encoders and present some result on movie scene content rating. However, the authors do not consider multi-modal and multi-label setting and focus on single-label classification using only video modality as input. A summary of the video-based movie content rating papers can be seen in Table \ref{table:vid_mov_rating}.

\section{Conclusion}
In this paper, related works are summarized for content action classification, multi-modal learning, movie genre classification, and sensitive content detection in the context of content moderation and movie content rating. The project page is available at \href{https://github.com/fcakyon/content-moderation-deep-learning}{https://github.com/fcakyon/content-moderation-deep-learning} and will be updated as new papers are published.

\printbibliography

\end{document}